\documentclass[10pt,twocolumn,letterpaper]{article}

\usepackage[pagenumbers]{cvpr} 

\usepackage{booktabs}
\usepackage{multirow}
\usepackage[table]{xcolor}

\definecolor{cvprblue}{rgb}{0.21,0.49,0.74}
\usepackage[pagebackref,breaklinks,colorlinks,allcolors=cvprblue]{hyperref}
\usepackage{algpseudocode}
\usepackage{algorithm}

\def\paperID{} 
\def\confName{CVPR}
\def\confYear{2026}

\title{E-GRPO: High Entropy Steps Drive Effective Reinforcement Learning \\ for Flow Models}

\author{Shengjun Zhang, Zhang Zhang, Chensheng Dai, Yueqi Duan$^{\dag}$\\
Tsinghua University \\
{\tt \small \{zhangsj23, z-z23\}@mails.tsinghua.edu.cn, duanyueqi@tsinghua.edu.cn}}

\begin{document}
\maketitle
\begin{abstract}
Recent reinforcement learning has enhanced the flow matching models on human preference alignment. 
While stochastic sampling enables the exploration of denoising directions, existing methods which optimize over multiple denoising steps suffer from sparse and ambiguous reward signals.
We observe that the high entropy steps enable more efficient and effective exploration while the low entropy steps result in undistinguished roll-outs.
To this end, we propose E-GRPO, an entropy aware Group Relative Policy Optimization to increase the entropy of SDE sampling steps. 
Since the integration of stochastic differential equations suffer from ambiguous reward signals due to stochasticity from multiple steps, we specifically merge consecutive low entropy steps to formulate one high entropy step for SDE sampling, while applying ODE sampling on other steps. 
Building upon this, we introduce multi-step group normalized advantage, which computes group-relative advantages within samples sharing the same consolidated SDE denoising step.
Experimental results on different reward settings have demonstrated the effectiveness of our methods.
Our code is available at \url{https://github.com/shengjun-zhang/VisualGRPO}.
\vspace{-0.8cm}
\end{abstract}    
\section{Introduction}
\label{sec:intro}

\begin{figure*}
    \centering
    \includegraphics[width=\linewidth]{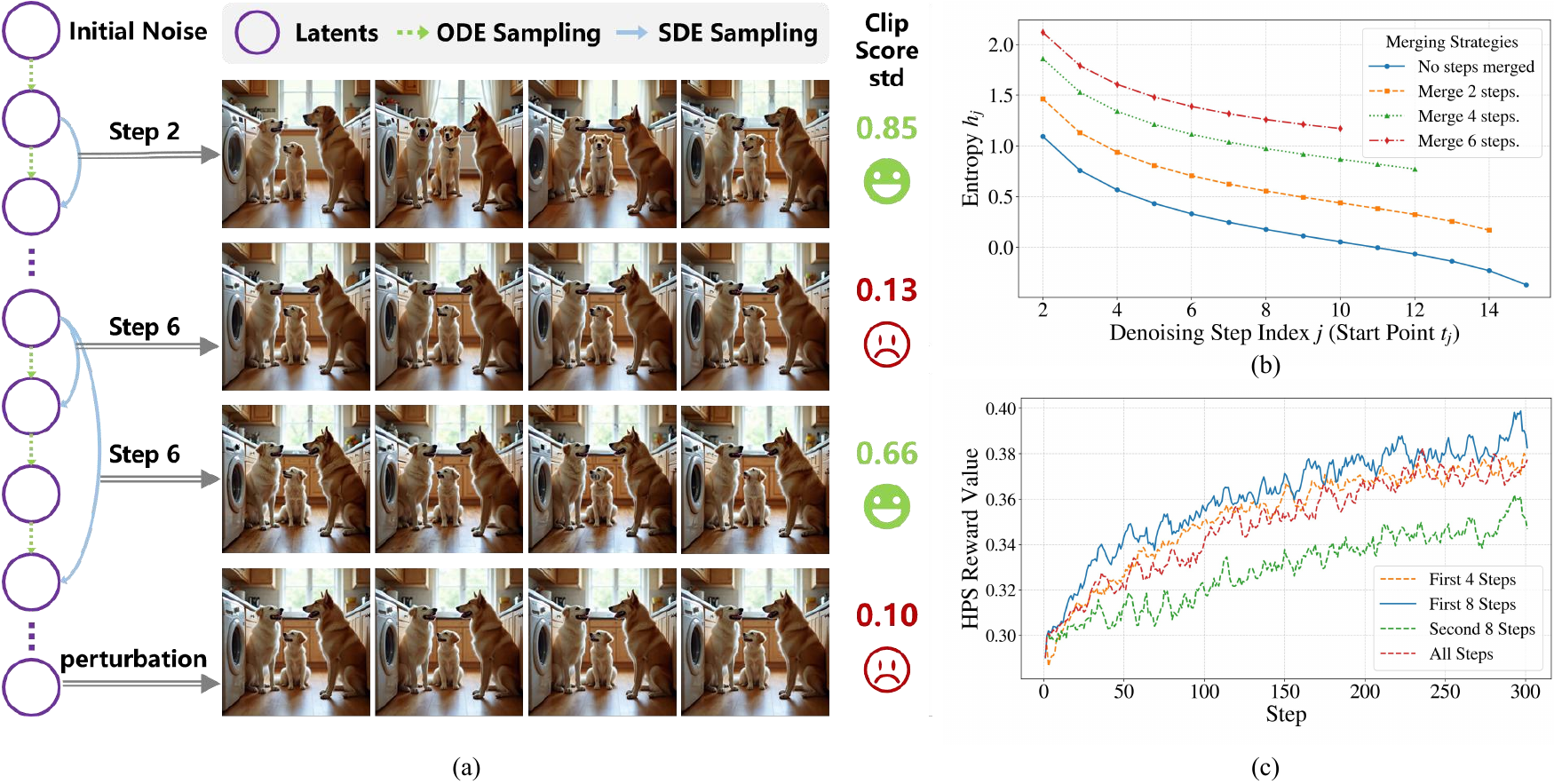}
    \vspace{-0.3cm}
    \caption{\textbf{The influence of entropy for sampling results.} (a) We visualize the generation images with different SDE sampling strategy, including one-step SDE on step 2, one-step SDE on step 6, and merged-step SDE on step 6. We also report the variance of clip score for generated images. Samples from the initial steps and merged steps share higher differences, while posterior steps generate undistinguishable samples, whose variance is similar to small perturbation on original images. (b) We report the entropy of SDE sampling on each timestep with different merged steps. More merged steps indicate higher entropy and larger exploration scope in RL training. (c) We visualize the training reward curves on models trained on all timesteps, the first half timesteps, and the second half timesteps.}
    \label{fig:teaser}
    \vspace{-0.3cm}
\end{figure*}

Recent advances in generative models have significantly propelled the field of visual content creation, enabling a wide array of applications ranging from artistic design and entertainment to medical imaging and virtual reality. 
State-of-the-art diffusion models~\cite{DDPM2020NIPS, DDIM2020arXiv, LDM2022CVPR} and flow-based approaches~\cite{FlowMatching2022arXiv, RectifiedFlow2022arXiv} have achieved remarkable fidelity in generating high-quality images and videos~\cite{SDXL2023arXiv, RFT2024ICML, SVD2023arXiv}. 

In large language models, reinforcement learning has demonstrated its effectiveness on the alignment with human preferences, including Proximal Policy Optimization (PPO)~\cite{PPO2017arXiv}, Direct Policy Optimization (DPO)~\cite{DPO2023NIPS}, and Group Relative Policy Optimization (GRPO)~\cite{GRPO2024arXiv}.
Thus, reinforcement learning from human feedback (RLHF)~\cite{DiffusionRL2023arXiv, Dpok2023NIPS} has been employed in post-training stages for visual generation.
Since GRPO simplifies the architecture by eliminating the value network, using intra-group relative rewards to compute advantages directly, recent works~\cite{DanceGRPO2025arXiv, Flowgrpo2025arXiv} integrate this into flow models with stochastic differential equations (SDE).
To enhance the efficiency of sampling, some methods~\cite{Mixgrpo2025arXiv, Tempflowgrpo2025arXiv, Chunkgrpo2025arXiv, G2RPO2025arXiv} introduce a mixture of SDE sampling and ODE sampling, while others~\cite{Branchgrpo2025arXiv, DynamicTreeGRPO2025arXiv} propose a tree-based structure for less sampling steps.

Despite these advancements, existing GRPO-based methods apply policy optimization across multiple denoising timesteps, resulting in sparse and ambiguous reward signals that hinder effective alignment. 
We observe that only high-entropy timesteps contribute meaningfully to training dynamics. 
As shown in Figure~\ref{fig:teaser} (b), stochastic exploration via SDE at timesteps with higher noise level possess larger entropy.
We visualize the generated images under different circumstances, where high-entropy timesteps yield diverse images with distinguishable reward variations, while low-entropy timesteps produce less reward differences, which are similar to those induced by adding 10\% random noise to the final image.
This phenomenon implies that reward models struggle to discern subtle trajectory deviations in low-entropy regimes.
Furthermore, we implement GRPO with SDE sampling on four settings: (i) the first 4 timesteps, (ii) the first 8 timesteps, (iii) the second 8 timesteps, and (iv) all 16 steps.
Notably, optimization on the first half timesteps performs even better than on all timesteps, which indicates that the second half timesteps are largely uninformative.

To address this limitation, we propose E-GRPO, an entropy-aware SDE sampling strategy for more effective exploration during GRPO training. 
An intuitive approach would be to employ multi-step continuous SDE sampling to broaden exploration.
However, this introduces cumulative stochasticity results in ambiguous reward attribution across steps, so that a beneficial exploration in one step may be penalized due to suboptimal downstream trajectory deviations, leading to optimization in the opposing direction.
Instead, we consolidate multiple low-entropy SDE steps into a single effective SDE step while keeping the remaining steps deterministic as ODE sampling, thereby preserving high-entropy exploration only where informative and ensuring reliable reward attribution. 
Building upon this, we introduce multi-step group normalized advantage, which computes group-relative advantages within samples sharing the same consolidated SDE step. 
This mechanism provides dense and trustworthy reward signals, enhancing the alignment of generative trajectories with human preferences.

We conduct experiments on both single-reward settings and multi-reward settings and evaluation on in-domain and out-of-domain matrices.
Experimental results demonstrate the effectiveness and efficiency of our method.
Our main contribution can be summarized as follows:
\begin{enumerate}
    \item We provide a comprehensive entropy-based analysis of denoising timesteps in GRPO training process, revealing that effective alignment can be achieved by optimizing exclusively at high-entropy steps.
    \item We propose E-GRPO, an entropy-aware SDE sampling strategy for GRPO training of flow models, which consolidates multiple low-entropy steps into a single high-entropy SDE step, thereby expanding meaningful exploration while eliminating reward attribution ambiguity.
    \item We conduct extensive experiments under both single-reward and multi-reward settings, clearly demonstrating that E-GRPO consistently outperforms prior methods, validating the efficacy and robustness of targeted, entropy-guided stochastic optimization.
\end{enumerate}

\section{Related Works}

\textbf{RL Alignment for Image Generation.}
Reinforcement Learning from Human Feedback (RLHF)~\cite{RLHF2022NIPS, Qwen252025arXiv} and Reinforcement Learning with Verifiable Rewards (RLVR)~\cite{TULU2025COLM} have emerged as powerful paradigms for aligning large language models (LLMs) with human preferences~\cite{Qwen252025arXiv, RLHF-22024arXiv, LMRLGym2023arXiv}.
Early frameworks based on Proximal Policy Optimization (PPO)~\cite{PPO2017arXiv} rely on a value model to guide policy updates, whereas recent approaches such as Group Relative Policy Optimization (GRPO)~\cite{GRPO2024arXiv, DeepseekR12025arXiv, OpanAIo12025arXiv2024} achieve greater stability and efficiency by leveraging relative group-wise comparisons instead of absolute rewards.
These advancements in language alignment have inspired increasing interest in transferring RL techniques to align visual generative models with human preferences.
In the visual generation domain, diffusion~\cite{DDPM2020NIPS, DDIM2020arXiv} and flow matching models~\cite{FlowMatching2022arXiv, RectifiedFlow2022arXiv, DiT2023NIPS} have demonstrated strong generative capabilities through iterative denoising processes~\cite{LDM2022CVPR, SDXL2023arXiv}.
To enhance alignment with human feedback, recent studies have adapted RLHF to these models.
Diffusion-DPO~\cite{DiffusionDPO2024CVPR}, and D3PO~\cite{D3PO2024CVPR} extend Direct Preference Optimization (DPO)~\cite{DPO2023NIPS} to diffusion models. However, these methods suffer from distribution shifting because no new samples are generated during the training process. While DanceGRPO~\cite{DanceGRPO2025arXiv} and Flow-GRPO~\cite{Flowgrpo2025arXiv} reformulate deterministic ODE-based sampling into stochastic SDE trajectories, enabling GRPO-style policy updates in visual domains.
Building upon this foundation, Granular-GRPO~\cite{G2RPO2025arXiv} refines timestep granularity for more precise and dense credit assignment across denoising steps, and TempFlow-GRPO~\cite{Tempflowgrpo2025arXiv} introduces temporally-aware weighting to alleviate the limitations of uniform optimization across timesteps.
MixGRPO~\cite{Mixgrpo2025arXiv} further improves training efficiency through a hybrid ODE–SDE sampling mechanism, while BranchGRPO~\cite{Branchgrpo2025arXiv} enhances exploration efficiency via branching rollouts and structured pruning.
Despite these advancements, existing GRPO frameworks for flow models typically optimize uniformly across all timesteps, overlooking the heterogeneity of exploration potential during the denoising process and suffering from sparse or noisy reward signals.
Our work addresses these challenges by leveraging step-wise entropy as a measure of exploration capacity, enabling optimization on high entropy steps to improve both stability and efficiency.

\noindent\textbf{Entropy-Guided Exploration and Alignment.}
Early work in reinforcement learning (RL) has recognized the importance of entropy as a mechanism for promoting effective exploration. In particular, strategies such as policy entropy regularization have been widely used to stabilize learning and encourage diverse behavior~\cite{Understanding2018PMLR}. For example, Soft Actor-Critic (SAC)~\cite{SAC2018arXiv} explicitly maximizes the expected reward while also maximizing policy entropy, resulting in more robust and sample-efficient exploration.
More recently, entropy-based insights have been applied to large language models (LLMs) in the context of reinforcement learning for reasoning. Study shows that a small fraction of high-entropy tokens disproportionately drives policy improvement, highlighting the significance of token-level uncertainty in guiding exploration~\cite{beyong80202025arXiv}. Complementary work further formalizes entropy as a lens for understanding exploration dynamics, demonstrating that high-entropy regions correspond to critical decision points that are most informative for learning~\cite{ReasonEntropy2025arXiv}. 
Inspired by these findings, we investigate whether similar entropy-driven patterns arise in flow matching models, and propose an entropy-aware GRPO framework that prioritizes informative denoising steps, leading to more efficient and stable alignment with human preferences.
\section{Methods}

\subsection{Preliminary}
To enable exploration in reinforcement learning, flow-based Group Relative Policy Optimization (GRPO) converts deterministic ODE sampling:
\begin{equation}
\mathrm{d}\mathbf{x}_t = \mathbf{v}_\theta(\mathbf{x}_t, t) \, \mathrm{d}t
\end{equation}
into an equivalent SDE:
\begin{align}
\mathbf{x}_{t+\Delta t} = & \mathbf{x}_t + \left[ \mathbf{v}_\theta(\mathbf{x}_t, t) + \frac{\sigma_t^2}{2t} \big(\mathbf{x}_t + (1-t)\mathbf{v}_\theta(\mathbf{x}_t, t)\big) \right] \Delta t \nonumber
\\ &+ \sigma_t \sqrt{\Delta t} \, \boldsymbol{\epsilon},
\end{align}
with $\boldsymbol{\epsilon} \sim \mathcal{N}(0, I)$ and $\sigma_t = a \sqrt{\frac{t}{1-t}}$.
With SDE sampling, flow-based GRPO integrates online reinforcement learning into flow matching models by framing the reverse sampling as a Markov Decision Process (MDP) with states $\mathbf{s}_t = (\mathbf{x}_t, t)$, actions $\mathbf{a}_t = \mathbf{x}_{t-1} \sim \pi_\theta(\cdot | \mathbf{s}_t)$, and terminal rewards $R(\mathbf{x}_0, c)$ for prompt $c$. The policy optimizes 
\begin{equation}
J_{\text{Flow-GRPO}}(\theta) = \mathbb{E}_{c \sim \mathcal{C},\ \{\mathbf{x}^{(i)}\}_{i=1}^G \sim \pi_{\theta_{\text{old}}}(\cdot|c)} \left[ f(r, A, \theta, \epsilon) \right]. \nonumber
\end{equation}
The clipped surrogate objective $f(r, A, \theta, \epsilon)$ is defined as:
\begin{align}
\frac{1}{G} \sum_{i=1}^G \frac{1}{T} \sum_{t=0}^{T-1} \Big[ \min\!\big(r_t^{(i)} A^{(i)}, \text{clip}(r_t^{(i)}, 1-\epsilon, 1+\epsilon) A^{(i)}\big) \notag 
\Big], \nonumber
\end{align}
where 
$r_t^{(i)}(\theta) = \frac{p_\theta(\mathbf{x}_{t-1}^{(i)} \mid \mathbf{x}_t^{(i)}, c)}{p_{\theta_{\text{old}}}(\mathbf{x}_{t-1}^{(i)} \mid \mathbf{x}_t^{(i)}, c)}$, and $p_\theta(\mathbf{x}_{t-1}^{(i)} \mid \mathbf{x}_t^{(i)}, c)$ is the policy function for output $\mathbf{x}^{(i)}$ at timestep $t-1$.
The group-normalized advantages $A^{(i)}$ is formulated as:
\begin{equation}
A^{(i)} = \frac{R(\mathbf{x}_0^{(i)}, c) - \text{mean}\{R(\mathbf{x}_0^{(j)}, c)\}_{j=1}^G}{\text{std}\{R(\mathbf{x}_0^{(j)}, c)\}_{j=1}^G}.
\end{equation}
Following the practices of previous methods~\cite{DanceGRPO2025arXiv, Mixgrpo2025arXiv}, the KL-regularization item is omitted in the objective function.

\begin{figure}
    \centering
    \includegraphics[width=0.9\linewidth]{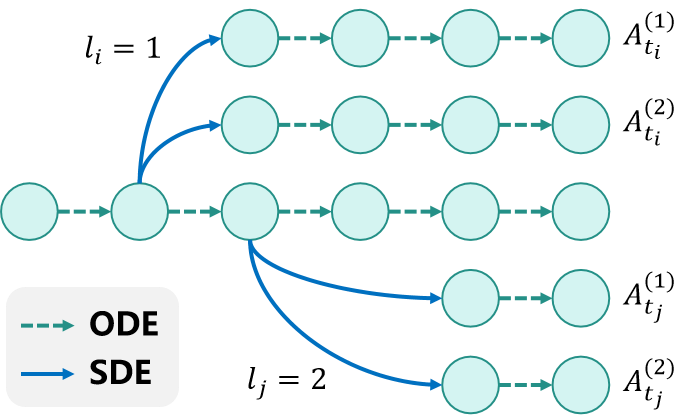}
    \caption{\textbf{E-GRPO sampling strategy.} First, we generate a set of anchor noise latents corresponding to different timesteps. For each active SDE timestep $t_i$, merged steps $\mathcal{T}_i$ is selected based on entropy analysis. We generate a group of results based on the specific SDE sampling of merged steps, and compute the advantage within each group.}
    \label{fig:pipeline}
\end{figure}

\subsection{Entropy Analysis}
In flow-based Group Relative Policy Optimization, the reverse sampling process from an SDE is framed as a Markov Decision Process (MDP). To derive the entropy of the reverse SDE step, we start from the given forward SDE and apply Bayes' theorem.
The forward SDE is given by:
\begin{align}
\mathbf{x}_{t+\Delta t} = \mathbf{x}_t + \mu_{\theta}(\mathbf{x}_t, t) \Delta t + \sigma_t \sqrt{\Delta t}\boldsymbol{\epsilon}, \label{eq:forward SDE}
\end{align}
where $\boldsymbol{\epsilon} \sim \mathcal{N}(0, I)$ injects stochasticity, and the drift term is:
\begin{align}
\mu_{\theta}(\mathbf{x}_t, t) = \mathbf{v}_\theta(\mathbf{x}_t, t) + \frac{\sigma_t^2}{2t} \big(\mathbf{x}_t + (1-t)\mathbf{v}_\theta(\mathbf{x}_t, t)\big). \nonumber
\end{align}
The transition probability of the forward SDE is a Gaussian distribution:
\begin{align}
p_{\text{f}}(\mathbf{x}_{t+\Delta t} \mid \mathbf{x}_t) = \mathcal{N}\left( \mathbf{x}_{t+\Delta t} \mid \mathbf{x}_t + \mu_{\theta}(\mathbf{x}_t, t) \Delta t,\ \sigma_t^2 \Delta t \, I \right). \nonumber
\end{align}

\noindent\textbf{Reverse SDE via Bayes' Theorem.}
The reverse transition probability $p_{\text{r}}(\mathbf{x}_t \mid \mathbf{x}_{t+\Delta t})$, which corresponds to the policy $\pi_\theta$ in GRPO, can be derived using Bayes' theorem:
\begin{align}
p_{\text{r}}(\mathbf{x}_t \mid \mathbf{x}_{t+\Delta t}) = \frac{p_{\text{f}}(\mathbf{x}_{t+\Delta t} \mid \mathbf{x}_t) \, p(\mathbf{x}_t)}{p(\mathbf{x}_{t+\Delta t})}. \nonumber
\end{align}
For a Gaussian process, the reverse transition is also a Gaussian distribution:
\begin{align}
p_{\text{r}}(\mathbf{x}_t \mid \mathbf{x}_{t+\Delta t}) = \mathcal{N}\left( \mathbf{x}_t \mid \tilde{\mu}_{\theta}(\mathbf{x}_{t+\Delta t}, t),\ \tilde{\sigma}_t^2 \Delta t \, I \right), \nonumber
\end{align}
where $\tilde{\mu}_{\theta}$ is the reverse drift and $\tilde{\sigma}_t$ is the reverse diffusion coefficient.
For linear Gaussian SDEs, the diffusion coefficient is the same in both directions when the process is time-reversible, where $\tilde{\sigma}_t = \sigma_t$.
For the reverse drift $\tilde{\mu}_{\theta}$, the log of the forward transition probability is:
\begin{align}
\log p_{\text{f}} = & -\frac{1}{2} \log \det(2\pi \sigma_t^2 \Delta t I) \nonumber \\
& -\frac{1}{2\sigma_t^2 \Delta t} \left\| \mathbf{x}_{t+\Delta t} - \mathbf{x}_t - \mu_{\theta}(\mathbf{x}_t, t) \Delta t \right\|^2 \nonumber
\end{align}
Taking the derivative with respect to $\mathbf{x}_t$, we find the reverse drift is formulated as:
\begin{align}
\tilde{\mu}_{\theta}(\mathbf{x}_{t+\Delta t}, t) = \mathbf{x}_{t+\Delta t} - \mu_{\theta}(\mathbf{x}_t, t) \Delta t + \sigma_t^2 \Delta t \, \nabla_{\mathbf{x}_t} \log p(\mathbf{x}_t). \nonumber
\end{align}

\noindent\textbf{Entropy of the Reverse SDE Step.}
The entropy of a multivariate Gaussian distribution $\mathcal{N}(\boldsymbol{\mu}, \boldsymbol{\Sigma})$ is given by:
\begin{align}
h(\mathbf{y}) = \frac{d}{2} \log\left( (2\pi e)^d \det(\boldsymbol{\Sigma}) \right)
\end{align}
where $d$ is the dimension of the random variable $\mathbf{y}$.
For the reverse SDE step, the covariance matrix is given by:
\begin{equation}
    \boldsymbol{\Sigma}_{\text{r}} = \sigma_t^2 \Delta t \, I. \label{eq:reserve convariance}
\end{equation}
The determinant of this diagonal matrix is $(\sigma_t^2 \Delta t)^d$. Substituting this into the entropy formula:
\begin{align}
h(t) &= \frac{1}{2} \log\left( (2\pi e)^d (\sigma_t^2 \Delta t)^d \right) \\
&= \frac{1}{2} \left[ d \log(2\pi e) + d \log(\sigma_t^2 \Delta t) \right] \\
&= \frac{d}{2} \log\left( 2\pi e \sigma_t^2 \Delta t \right)
\end{align}
Substituting $\sigma_t = a \sqrt{\frac{t}{1-t}}$, we get:
\begin{align}
h(t) = \frac{d}{2} \log\left( 2\pi e \cdot a^2 \cdot \frac{t}{1-t} \cdot \Delta t \right) \label{eq:entropy}
\end{align}

\begin{figure}
    \centering
    \includegraphics[width=0.9\linewidth]{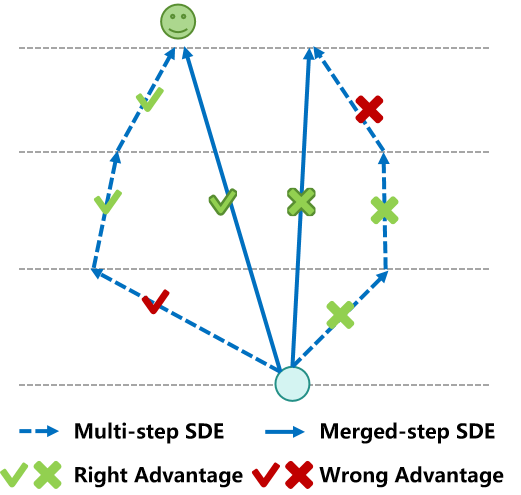}
    \caption{\textbf{Ambiguous reward signal.} For consecutive multi-step SDE sampling, the advantage is corresponding to multiple timesteps, which may results in wrong optimization direction on the specific timestep. Our merged-step SDE sampling not only enlarges the exploration scope, but also eliminate ambiguous reward by aligning the final advantage to one merged SDE step.
    }
    \label{fig:merge_step}
\end{figure}

\subsection{Entropy-aware GRPO}
To address the sparse and ambiguous reward attribution of uniform optimization across timesteps, we propose an entropy-aware GRPO (E-GRPO) framework, which integrates an entropy-driven step merging strategy and multi-step group normalized advantage estimation. The core design prioritizes meaningful exploration by consolidating low-entropy SDE steps into informative sampling events.

\subsubsection{Entropy-Driven Step Merging Strategy}~\label{sec:entropy driven merge}
Given a sequence of denoising timesteps \(\{t_T, ..., t_1, t_0\}\), the relation of timesteps and the entropy is formulated as:
\begin{align}
   e^{h(t_k)}\propto \dfrac{t_k}{1-t_k}(t_{k}-t_{k-1}). \label{eq:exp entropy}
\end{align}
Practically, flow models adjust time shift to balance quality and efficiency:
\begin{equation}
    t_k=\frac{s\hat{t}_k}{1+(s-1)\hat{t}_k}, \nonumber
\end{equation}
where $\hat{t}_k=\frac{k}{T}$.
Substituting this into (\ref{eq:exp entropy}), we get:
\begin{equation}
    e^{h(t_k)}\propto \dfrac{s^2 T k}{(T-k)[T+(s-1)k][T+(s-1)(k-1)]}. \nonumber
\end{equation}

We define an adaptive entropy threshold $\tau$ to classify timesteps into high-entropy ones $\{t_{T},\cdots,t_{M+1}\}$ with $e^{h(t_k)}\geq\tau$ and low-entropy ones $\{t_{M},\cdots, t_0\}$ with $e^{h(t_k)}<\tau$. 
For a low-entropy timestep $t_m$, we can introduce multi-step SDE sampling on consecutive timesteps $\{t_{m}, ..., t_{m-l}\}$.
As shown in Figure~\ref{fig:merge_step}, this introduces cumulative stochasticity results in ambiguous reward attribution across steps, so that a beneficial exploration in one step may be penalized due to suboptimal downstream trajectory deviations, leading to optimization in the opposing direction. 
Thus, we consolidate consecutive timesteps into a single equivalent SDE step to eliminate ambiguous reward signals.
Merging \(l\) consecutive low-entropy SDE steps requires preserving the total diffusion effect while reducing step count.
For the consolidated timesteps, the time interval is $\Delta t=t_{m}-t_{m-l}$.
Substituting $\Delta t$ to (\ref{eq:forward SDE}), we have:
\begin{equation}
    \mathbf{x}_{t_{m-l}} = \mathbf{x}_{t_m} + \mu_{\theta}(\mathbf{x}_{t_m}, t_m)(t_{m}-t_{m-l}) + \sigma_t \sqrt{t_{m}-t_{m-l}} \, \boldsymbol{\epsilon}. \nonumber
\end{equation}
According to (\ref{eq:reserve convariance}), the reverse SDE step is formulated as:
\begin{equation}
    \boldsymbol{\Sigma} = \sigma_t^2 (t_{m}-t_{m-l}) \, I. \nonumber
\end{equation}
Thus, the entropy for the merged timestep is given by:
\begin{align}
    e^{h(t_k)} & \propto \dfrac{t_m}{1-t_m}(t_{m}-t_{m-l}) \nonumber \\
    & \propto \dfrac{s^2 T m l}{(T-m)[T+(s-1)m][T+(s-1)(m-l)]}, \nonumber
\end{align}
where $e^{h(t_k)}$ is an increasing function of $l$.

Instead of using a uniform $l$ for all low-entropy timesteps, we propose an adaptive strategy to select an optimal $l$ for each low-entropy timestep, where $l$ is determined such that the entropy of the merged step just exceeds the threshold $\tau$. 
This design avoids excessively large entropy of a single merged step, which would make it difficult to find a proper optimization direction under limited exploration attempts.
The adaptive selection of $l$ ensures that each merged step maintains a moderate entropy level—sufficient to retain meaningful exploration signals while preventing the entropy from becoming too high to guide effective optimization. 
By aligning the merged entropy with the predefined threshold, we balance the efficiency gain from step merging and the reliability of reward-guided exploration.

\begin{algorithm}[t]
\caption{Entropy-aware GRPO (E-GRPO)}\label{alg:e-grpo}
\renewcommand{\algorithmicrequire}{\textbf{Input:}}
\renewcommand{\algorithmicensure}{\textbf{Output:}}
\begin{algorithmic}[1]
\Require{Initial policy $\theta_{\text{old}}$, prompt set $\mathcal{C}$, total timesteps $T$, active SDE sampling timesteps $\{t_T, \cdots, t_N\}$, merging step count $\{l_T, \cdots, l_N\}$, clipping coefficient $\epsilon$, trajectory count $\{G^{(T)},\cdots, G^{(N)}\}$}
\Ensure{Optimized policy $\theta$}
\For{iteration = 1 to $K$ (total iterations)}
    \For{$c \sim \mathcal{C}$ (sample prompt)}
        \For{$N \leq n \leq T$}
            \State $\mathcal{T}_n\leftarrow\{t_n, t_{n-1}, \cdots, t_{n-l_n}\}$
            \State Generate $G^{(n)}$ trajectories with $\mathcal{T}_n$
            \State Compute rewards $\{R(\mathbf{x}_{0,t_n}^{(j)}, c)\}_{j=1}^{G^{(n)}}$ 
            \State $A_{t_n}^{(i)}\leftarrow\frac{R(\mathbf{x}_{0,t_n}^{(i)}, c) - \text{mean}\{R(\mathbf{x}_{0,t_n}^{(j)}, c)\}_{j=1}^{G^{(n)}}}{\text{std}\{R(\mathbf{x}_{0,t_n}^{(j)}, c)\}_{j=1}^{G^{(n)}}}$
            \State $r_{t_n}^{(i)}(\theta)\leftarrow\frac{p_\theta(\mathbf{x}_{t_{n-l_n}}^{(i)} \mid \mathbf{x}_{t_n}^{(i)}, c)}{p_{\theta_{\text{old}}}(\mathbf{x}_{t_{n-l_n}}^{(i)} \mid \mathbf{x}_{t_n}^{(i)}, c)}$
        \EndFor
        \State Construct clipped surrogate: $f(r,A,\theta,\epsilon)$
        \State Update $\theta$ by minimizing $J_{\text{E-GRPO}}(\theta)$
        \State Set $\theta_{\text{old}} \leftarrow \theta$ 
    \EndFor
\EndFor
\Return Optimized policy $\theta$
\end{algorithmic}
\end{algorithm}

\begin{table*}[ht]
  \centering
  \caption{\textbf{Evaluation Results. }Comparison between different methods. The best and second best results in each column are \textbf{bolded} and \underline{underline} respectively. }
  \label{tab:main_exp_table}
  
  
  \begin{tabular}{lcccccccc}
    \toprule
    \multirow{2}{*}{\textbf{Method}} & 
    \multicolumn{4}{c}{\textbf{Training Reward Model: HPS}} & 
    \multicolumn{4}{c}{\textbf{Training Reward Models: HPS\&CLIP}} \\
    
    \cmidrule(lr){2-5} \cmidrule(lr){6-9}
    
    & \textbf{HPS} & \textbf{CLIP} & \textbf{PickScore} & \textbf{ImageScore} 
    & \textbf{HPS} & \textbf{CLIP} & \textbf{PickScore} & \textbf{ImageScore} \\
    \midrule
    
    \rowcolor{gray!8}
    FLUX.1-dev~\cite{FLUX2024Github} & 0.311 & 0.388 & 0.231 & 1.089 & 0.311 & 0.388 & 0.231 & 1.089 \\
    \midrule
    DanceGRPO~\cite{DanceGRPO2025arXiv}    & 0.353 & \textbf{0.375} & 0.228 & 1.233 & 0.331 & 0.389 & 0.227 & 1.128 \\
    MixGRPO~\cite{Mixgrpo2025arXiv}      & 0.378 & 0.358 & 0.225 & 1.266 & 0.363 & 0.399 & 0.230 & 1.436 \\
    GranularGRPO~\cite{G2RPO2025arXiv} & \underline{0.385} & 0.355 & 0.229 & \underline{1.313} & \underline{0.377} & \underline{0.400} & \underline{0.236} & \underline{1.490} \\
    BranchGRPO~\cite{Branchgrpo2025arXiv}   & 0.358 & \underline{0.365} & \underline{0.231} & 1.311 & 0.342 & 0.384 & 0.230 & 1.243 \\
    TempFlowGRPO~\cite{Tempflowgrpo2025arXiv} & 0.382 & 0.357 & \underline{0.231} & 1.264   & 0.310 & 0.388 & 0.230 & 1.106 \\
    \midrule
    
    \textbf{Ours} & \textbf{0.391} & 0.355 & \textbf{0.232} & \textbf{1.324} & \textbf{0.382} & \textbf{0.401} & \textbf{0.237} & \textbf{1.494} \\
    
    \bottomrule
  \end{tabular}
\end{table*}

\subsubsection{Policy Optimization Objective}
To resolve reward attribution ambiguity, we extend GRPO's group normalization to merged steps by defining merge-grouped samples. We designate a set of active SDE timesteps $\{t_T, \cdots, t_N\}$, where each timestep $t_n$ (with $N \leq n \leq T$) is associated with a merging step count $l_n$ determined by the entropy-driven strategy in Section~\ref{sec:entropy driven merge}. For a given prompt $c$, we generate $G^n$ trajectories for each active timestep $t_n$, where all $G^n$ trajectories share the same consolidated merged timesteps $\mathcal{T}_n \triangleq \{t_n, t_{n-1}, \cdots, t_{n-l_n}\}$.

Within each merge group $\mathcal{T}_n$, we first compute the advantage estimates using the $G^n$ trajectories, ensuring reward signals are attributed consistently to the merged timesteps. 
The advantage of the $i$-th trajectory at state $\mathbf{x}_{t_n}$ estimated over the merge group $\mathcal{T}_n$ is given by: 
\begin{equation}
    A^{(i)}_{t_n}=\frac{R(\mathbf{x}_{0,t_n}^{(i)}, c) - \text{mean}\{R(\mathbf{x}_{0,t_n}^{(j)}, c)\}_{j=1}^{G_{n}}}{\text{std}\{R(\mathbf{x}_{0,t_n}^{(j)}, c)\}_{j=1}^{G_{n}}}, \label{eq:adavantage}
\end{equation}
where $\mathbf{x}_{0,t_n}^{(j)}$ denotes the $j$-th generated results with active SDE timestep $t_n$. The final clipped surrogate objective $f(r,A,\theta,\epsilon)$, adapted to merge-grouped samples, is then formulated as:
\begin{align}
    \frac{1}{\hat{T}} \sum_{n=N}^{T}\frac{1}{G^{(n)}} \sum_{i=1}^{G^{(n)}}\min\!\big(r^{(i)}_{t_n} A^{(i)}_{t_n}, \text{clip}(r_{t_n}^{(i)}, 1-\epsilon, 1+\epsilon) A^{(i)}_{t_n}\big), \nonumber
\end{align}
where $\hat{T}=T-N$ and the ratio $r^{(i)}_{t_n}$ is given by:
\begin{equation}
    r_{t_n}^{(i)}(\theta) = \frac{p_\theta(\mathbf{x}_{t_{n-l_n}}^{(i)} \mid \mathbf{x}_{t_n}^{(i)}, c)}{p_{\theta_{\text{old}}}(\mathbf{x}_{t_{n-l_n}}^{(i)} \mid \mathbf{x}_{t_n}^{(i)}, c)}. \label{eq:ratio}
\end{equation}
Building on the clipped surrogate objective of GRPO, E-GRPO restricts optimization to consolidated high-entropy steps. 
The objective function is modified to:
\begin{equation}
    J_{\text{E-GRPO}}(\theta) = \mathbb{E}_{c \sim \mathcal{C}, \{\mathbf{x}_{t_n}^{(i)}\}_{i=1}^{G^{(n)}} \sim \pi_{\theta_{\text{old}}}(\cdot|c),N\leq n\leq T}  f(r,A,\theta,\epsilon). \nonumber
\end{equation}

Our strategy is illustrated in Algorithm~\ref{alg:e-grpo}.
We first Compute \(h(t_k)\) for all timesteps using (\ref{eq:entropy}) and determine $\tau$.
Then, we cluster consecutive low-entropy steps $\mathcal{T}_i$ for timestep $t_i$.
We generate trajectories using ODE for other steps and consolidated SDE for merged steps.
Finally, we estimate advantages $A_{t_n}^{(i)}$ and ratio via (\ref{eq:adavantage}) and (\ref{eq:ratio}) so that reward signals are attributed consistently to the merged timesteps, and update $p_{\theta}$ by minimizing $J_{\text{E-GRPO}}(\theta)$.
\section{Experiments}

\begin{figure}[t]
    \centering
    \includegraphics[width=\columnwidth]{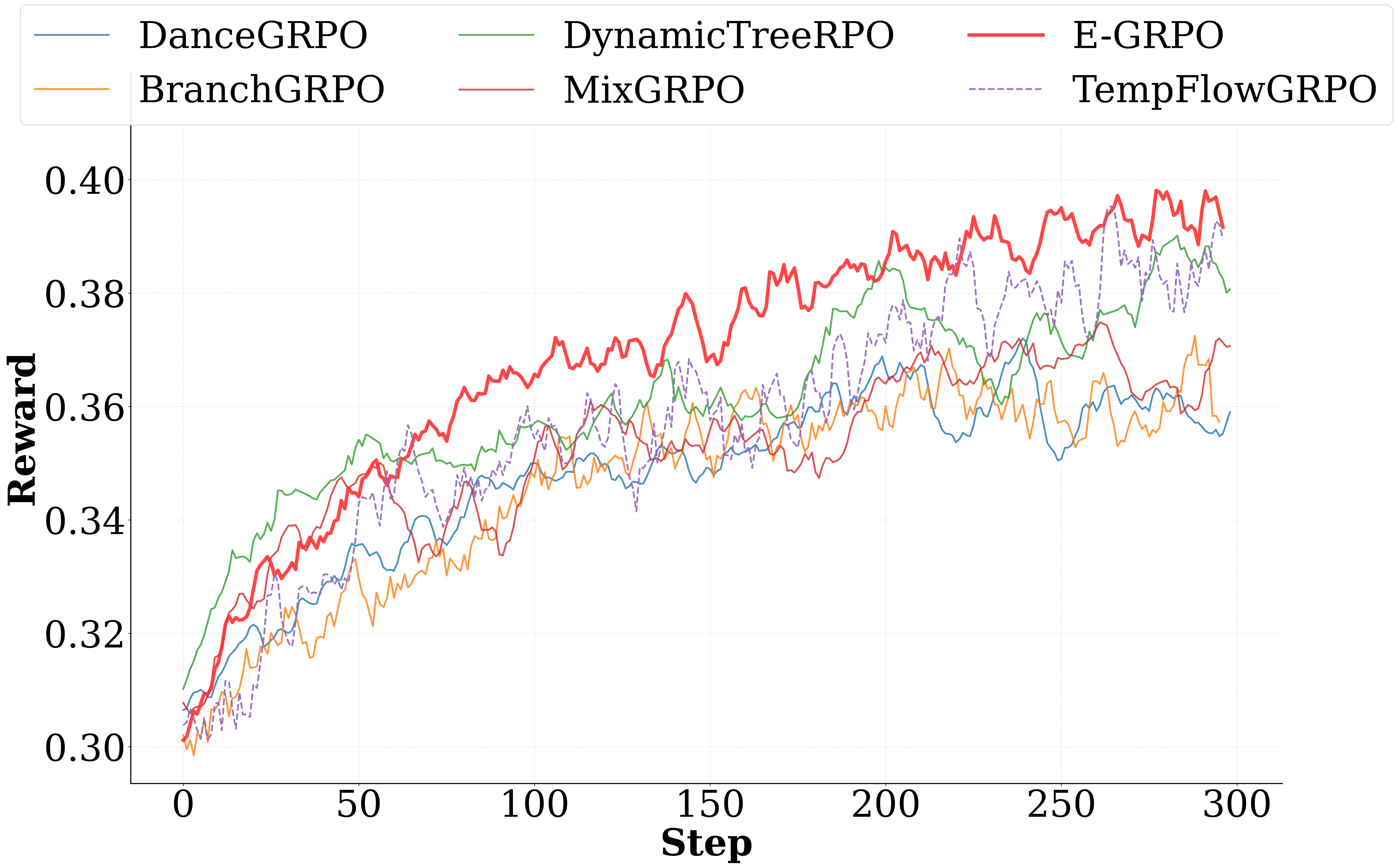}
    \caption{\textbf{Comparison of Training Reward Curves.} The reward curve of E-GRPO demonstrates faster and more stable improvement during training compared to baseline methods. This indicates that exploration guided by high-entropy steps can enhance learning efficiency while mitigating noise in the reward signal.}
    \label{fig:main_exp_reward_curve}
\end{figure}

\begin{figure*}[t]
    \centering
    \includegraphics[width=\textwidth]{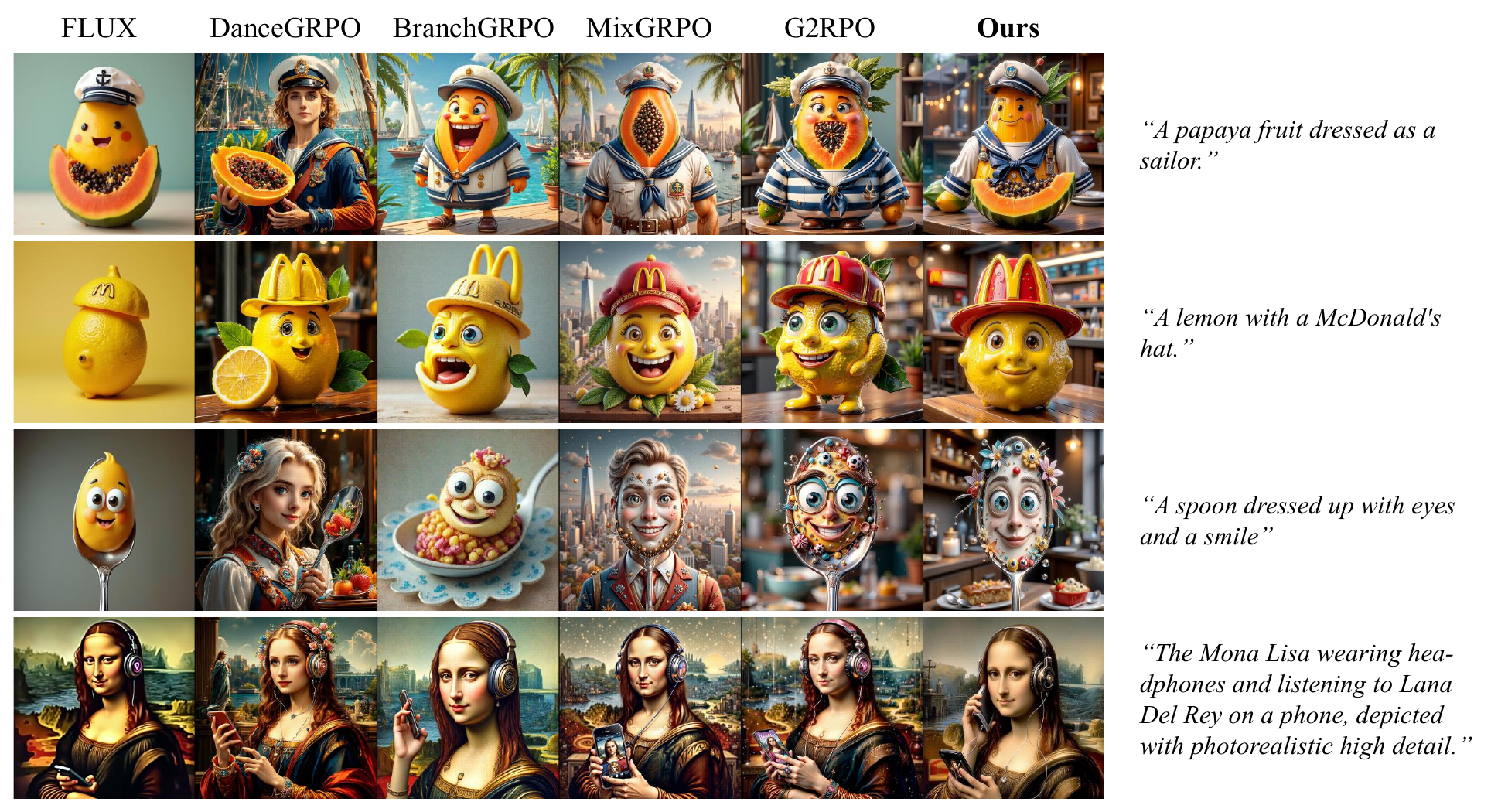}
    \caption{\textbf{Visualization Comparisons.} Comparison between E-GRPO with other baseline methods.  E-GRPO better integrates semantics and fine-grained details. }
    \label{fig:main_exp_visual}
\end{figure*}

\subsection{Experimental Settings}

\noindent \textbf{Dataset and Model. }We conduct our experiments on the HPD dataset~\cite{HPSv22023arXiv}, a large-scale dataset for human preference evaluation, containing approximately 103,000 text prompts for training and 400 prompts for testing.  
For our experiments, we adopt FLUX.1-dev~\cite{FLUX2024Github} as the backbone flow matching model, consistent with prior works such as DanceGRPO~\cite{DanceGRPO2025arXiv} and MixGRPO~\cite{Mixgrpo2025arXiv}.

\noindent \textbf{Evaluation Settings. }To assess alignment with human preferences, we employ several representative reward models, each capturing different aspects of generated images.
\textit{HPS-v2.1}~\cite{HPSv22023arXiv} and \textit{PickScore}~\cite{PickScore2023NIPS} are both trained on large-scale human preference data, thus reflecting human judgments of overall image quality and text–image consistency.
\textit{CLIP Score}~\cite{CLIP2021PMLR} primarily measures the semantic alignment between the generated image and the input prompt.
\textit{ImageReward}~\cite{ImageReward2023NIPS} focuses on the perceptual quality and aesthetic appeal of the image, providing a complementary perspective to preference-based metrics.

\noindent \textbf{Sampling Strategy. }  
Following DanceGRPO, images generated in training are sampled from the same initialized noise to form rollout groups.  
We set the total number of sampling steps to $T = 16$ and the parameter $a$ in the equation $\sigma_t = \displaystyle a\sqrt{\frac{t}{1 - t}}$ to $0.7$. During training, the entropy threshold $\tau$ for step merging is set as $2.2$.

\noindent \textbf{Training Details. }
We first train the model using only the HPS-v2.1 reward model to estimate the upper performance bound of our approach.
To enhance robustness and mitigate potential reward hacking, we further train the model with both HPS-v2.1 and CLIP as joint reward signals. Training is only conducted on the first half sampling steps. 

\noindent \textbf{Optimization Details. }
All experiments are performed on 8 $\times$ NVIDIA A800 GPUs with a batch size of 1.
We employ the AdamW optimizer with a learning rate $2\times 10^{-6}$ and a weight decay of $1 \times 10^{-4}$.
Mixed-precision training is enabled using the bfloat16 format.
The total number of training iterations is 300.

\subsection{Main Experiments}

As shown in Tab.~\ref{tab:main_exp_table}, we evaluate our method against several recent methods, including the baseline FLUX.1-dev~\cite{FLUX2024Github}, DanceGRPO~\cite{DanceGRPO2025arXiv}, MixGRPO~\cite{Mixgrpo2025arXiv}, BranchGRPO~\cite{Branchgrpo2025arXiv}, TempFlowGRPO~\cite{Tempflowgrpo2025arXiv} and GranularGRPO~\cite{G2RPO2025arXiv}.  
When trained with the single HPS-v2.1 reward, our method achieves a new state-of-the-art performance, surpassing DanceGRPO by 10.8\% on the HPS metric. 
This demonstrates that our entropy-guided exploration effectively identifies high-value denoising steps, leading to more precise and stable policy optimization.

However, as discussed in DanceGRPO~\cite{DanceGRPO2025arXiv}, training solely with HPS-v2.1 can lead to reward hacking, resulting in overly saturated visual results that do not align with genuine human preferences.  
To address this, we follow prior works~\cite{DanceGRPO2025arXiv, Mixgrpo2025arXiv, G2RPO2025arXiv} and adopt a joint reward scheme using both HPS-v2.1 and CLIP score as reward during training.  
Under this more robust multi-reward setting, our approach not only maintains its SOTA performance on the in-domain HPS metric but also achieves substantial improvements on out-of-domain metrics.  
In particular, compared with DanceGRPO, our method improves ImageReward by 32.4\% and PickScore by 4.4\%, highlighting that entropy-guided optimization promotes broader generalization across reward models and effectively mitigates reward hacking.

Figure~\ref{fig:main_exp_visual} presents qualitative comparisons among FLUX.1-dev, DanceGRPO, BranchGRPO, MixGRPO, G2RPO, and our proposed E-GRPO.
As shown in the first row (prompt: “A papaya fruit dressed as a sailor”), E-GRPO generates a composition that naturally integrates the papaya’s structure with human-like attire, yielding images of higher aesthetic quality and greater realism.
In contrast, baseline methods either misinterpret the prompt (e.g., DanceGRPO generates a person holding a papaya) or produce visually incoherent results (e.g., MixGRPO and G2RPO).
In the third row (prompt: “A spoon dressed up with eyes and a smile”), E-GRPO produces expressive and visually consistent humanized faces while preserving the metallic texture of the spoon, whereas other methods generate unrealistic facial blending or lose material fidelity.
These results highlight that E-GRPO achieves superior semantic grounding and visual coherence, leading to images that more faithfully reflect textual intent and human aesthetic preference.

Figure~\ref{fig:main_exp_reward_curve} illustrates the reward trajectories during training.  
Compared with prior work, our method exhibits faster early-stage reward growth and smoother convergence, achieving a higher final reward.  
This indicates that the entropy-guided step selection stabilizes optimization by focusing updates on the most informative denoising steps, improving both efficiency and reliability.

\subsection{Ablation Studies}

In order to evaluate the effectiveness of the proposed method, we conduct a series of ablation experiments to understand the design of E-GRPO. 

\noindent \textbf{Step Merging Strategies.}  
We evaluate several step merging strategies to verify the effectiveness of our entropy-based adaptive merging scheme during training.  
As shown in Tab.~\ref{tab:ablation_step_merge}, our method consistently outperforms the naive 2-step, 4-step, and 6-step merging baselines across almost all evaluation metrics, demonstrating both efficiency and robustness.  
Compared with fixed merging strategies that combine multiple steps regardless of their exploration level, our entropy-aware adaptive merging dynamically adjusts the merging behavior to maintain comparable exploration across steps, leading to more accurate and efficient optimization.

\begin{table}[t]
  \caption{\textbf{Comparison of Step Merging Strategies.}
  Quantitative results comparing different step merging strategies during training.  
  The proposed entropy-aware adaptive merging consistently achieves higher scores on HPS, CLIP, PickScore, and ImageScore, indicating better semantic alignment and generation quality. The best results in each column are \textbf{bolded}}.
  \label{tab:ablation_step_merge}
  \footnotesize
  
  \begin{tabular}{lcccc}
    \toprule

    \textbf{Merging Strategies} & \textbf{HPS} & \textbf{CLIP} & \textbf{PickScore} & \textbf{ImageScore} \\
    \midrule
    
    2-step & 0.382 & 0.290 & 0.232 & 1.223  \\
    4-step & 0.374 & 0.302 & 0.230 & 1.216  \\
    6-step & 0.337 & \textbf{0.372} & 0.226 & 1.298      \\
    \midrule
    \textbf{Adaptive} & \textbf{0.391} & 0.355 & \textbf{0.232} & \textbf{1.324} \\
    
    \bottomrule
  \end{tabular}
\end{table}

\noindent \textbf{Step Entropy Analysis.}
To validate the rationality of the entropy-based analysis and effectiveness of the proposed entropy-aware GRPO method, we conduct experiments by training models on different subsets of denoising steps.  
Specifically, we train separate models using (1) the first 4 steps, (2) the first 8 steps, (3) the last 8 steps, and (4) all steps.  
As shown in Fig.~\ref{fig:teaser}(c), training on the first 8 high-entropy steps achieves the best performance, followed by using the first 4 steps.  
In contrast, training on all steps yields similar results to the first 4-step case but with substantially higher computational cost.  
When the model is trained on the last 8 (low-entropy) steps, the performance drops dramatically.  
These results indicate that focusing training on early high-entropy steps is sufficient to achieve strong performance, while involving too many later low-entropy steps introduces unnecessary noise and inefficiency.  
Therefore, we adopt the first 8 denoising steps as our default training configuration. 
Tab.~\ref{tab:ablation_step_entropy} further provides quantitative results for different subsets of training steps.

\begin{table}[t]

  \caption{\textbf{Comparison of Different Training Denoising Steps.}
  Models trained on high-entropy (early) steps achieve higher alignment scores with lower computational cost, confirming that high-entropy steps contribute most to effective optimization. 
  The highest score in each column is \textbf{bolded}. 
  Note that the CLIP score shows an unexpected deviation, which is caused by training solely with the HPS reward, leading to a certain degree of reward hacking as discussed earlier.}

  \label{tab:ablation_step_entropy}
  \footnotesize 
  \begin{tabular}{lcccccccc}
    \toprule
    \textbf{Merging Strategies} & \textbf{HPS} & \textbf{CLIP} & \textbf{PickScore} & \textbf{ImageScore} \\
    \midrule
     First 4 Steps  & 0.370 & 0.348 & 0.231 & 1.252  \\
     First 8 Steps & \textbf{0.391} & 0.355 & \textbf{0.232} & \textbf{1.324} \\
     Second 8 Steps & 0.357 & \textbf{0.381} & 0.231 & 1.250  \\
     Full Steps     & 0.366 & 0.359 & 0.231 & 1.169  \\
    \bottomrule
  \end{tabular}
\end{table}

\section{Conclusion}
This work addresses the critical challenge of sparse and ambiguous reward signals in existing Group Relative Policy Optimization (GRPO)-based methods for flow models, which stem from uniform optimization across all denoising timesteps. 
Through entropy analysis, we reveal a key insight that high-entropy timesteps contribute meaningfully to effective exploration and human preference alignment, while low-entropy timesteps yield undistinguished rollouts that hinder reward discrimination. 
To tackle this limitation, we propose E-GRPO, an entropy-aware framework that integrates two core innovations, including an adaptive entropy-driven step merging strategy and multi-step group normalized advantage estimation. 
The step merging strategy consolidates consecutive low-entropy SDE steps into single high-entropy SDE steps, while retaining ODE sampling for other steps, eliminating reward attribution ambiguity caused by cumulative stochasticity.
The multi-step group normalized advantage ensures dense and reliable reward signals by computing relative advantages within samples sharing the same consolidated step. 
Extensive experiments on the HPD dataset with FLUX.1-dev as the backbone validate the efficacy of E-GRPO.

\noindent\textbf{Limitations and Future Works.} A critical bottleneck in advancing visual generative models lies in the design and alignment of reward signals. Rewards serve as the cornerstone of guiding reinforcement learning paradigms toward generating high-quality, human-preferred content, yet existing reward formulations often fail to fully align with nuanced human preferences—such as aesthetic appeal, semantic consistency, and contextual appropriateness. This misalignment not only leads to suboptimal generation outcomes but also renders models vulnerable to reward hacking: models may exploit loopholes in the reward function to maximize scores without genuinely meeting human expectations. As a result, the development of more robust and effective reward models remains an essential direction for future research in visual RL-driven generation.

\noindent\textbf{Acknowledgments.} This work was supported in part by the Beijing Natural Science Foundation under Grant L252011, and by the National Natural Science Foundation of China under Grant 62576185.

\clearpage
\maketitlesupplementary

\section{Ablation Study on the Entropy Threshold $\tau$}
\label{sec:tau_ablation}

\begin{table}[!b]
    \centering
    \caption{
        \textbf{Ablation study on the entropy threshold $\tau$.}
        Results are reported under the HPS reward setting.
        A threshold of $\tau=0$ corresponds to the baseline without step merging.
        Our default choice ($\tau=2.2$) achieves the overall best performance.
        Best results in each column are highlighted in \textbf{bold}.
    }
    \label{tab:tau_ablation}
    \begin{tabular}{@{}lcccc@{}}
        \toprule
        Threshold ($\tau$)  & HPS & CLIP & PickScore & ImageScore \\ \midrule
        0 (No Merging)      & 0.384 & 0.349 & 0.230 & 1.297 \\
        1.8                 & 0.383 & 0.352 & 0.232 & 1.293 \\
        2.0                 & 0.384 & 0.344 & 0.231 & 1.269 \\
        \textbf{2.2 (Ours)} & \textbf{0.391} & \textbf{0.355} & \textbf{0.233} & \textbf{1.324} \\
        2.6                 & 0.388 & 0.355 & 0.233 & 1.320 \\ 
        \bottomrule
    \end{tabular}
\end{table}

In our main paper, we introduce an adaptive entropy threshold $\tau$ to separate timesteps into high-entropy and low-entropy groups. During sampling, consecutive low-entropy steps are merged until their entropy reaches the threshold. This threshold serves as a critical hyperparameter in our entropy-driven step-merging strategy. To claim the effectiveness of the proposed method and assess its sensitivity to $\tau$, we conducted a series of experiments with different threshold values. Specifically, we trained E-GRPO with $\tau$ set to 0 (meaning all steps are treated equally and no merging ooccurs), 1.8, 2.0, 2.2 (our default setting), and 2.6 under the HPS reward configuration. The results are summarized in Table~\ref{tab:tau_ablation}.

As shown in Table~\ref{tab:tau_ablation}, the model behaves noticeably differently under varying threshold values. As $\tau$ increases, the achievable HPS score also improves, indicating the effectiveness of entropy as a guidance signal during training. However, when $\tau$ becomes excessively large, a long sequence of steps may be merged, occasionally combining steps that still contain useful entropy or gradient information. This leads to overly coarse updates and, consequently, a slight degradation in performance. Notably, our default choice of $\tau = 2.2$ strikes an effective balance between leveraging entropy for guidance and avoiding excessive merging, yielding the best overall performance in our experiments.

\section{Additional Visualizations}
\label{sec:more_visuals}

\subsection{More Quality Results}

\begin{figure*}[t] 
    \centering
    \includegraphics[width=\textwidth]{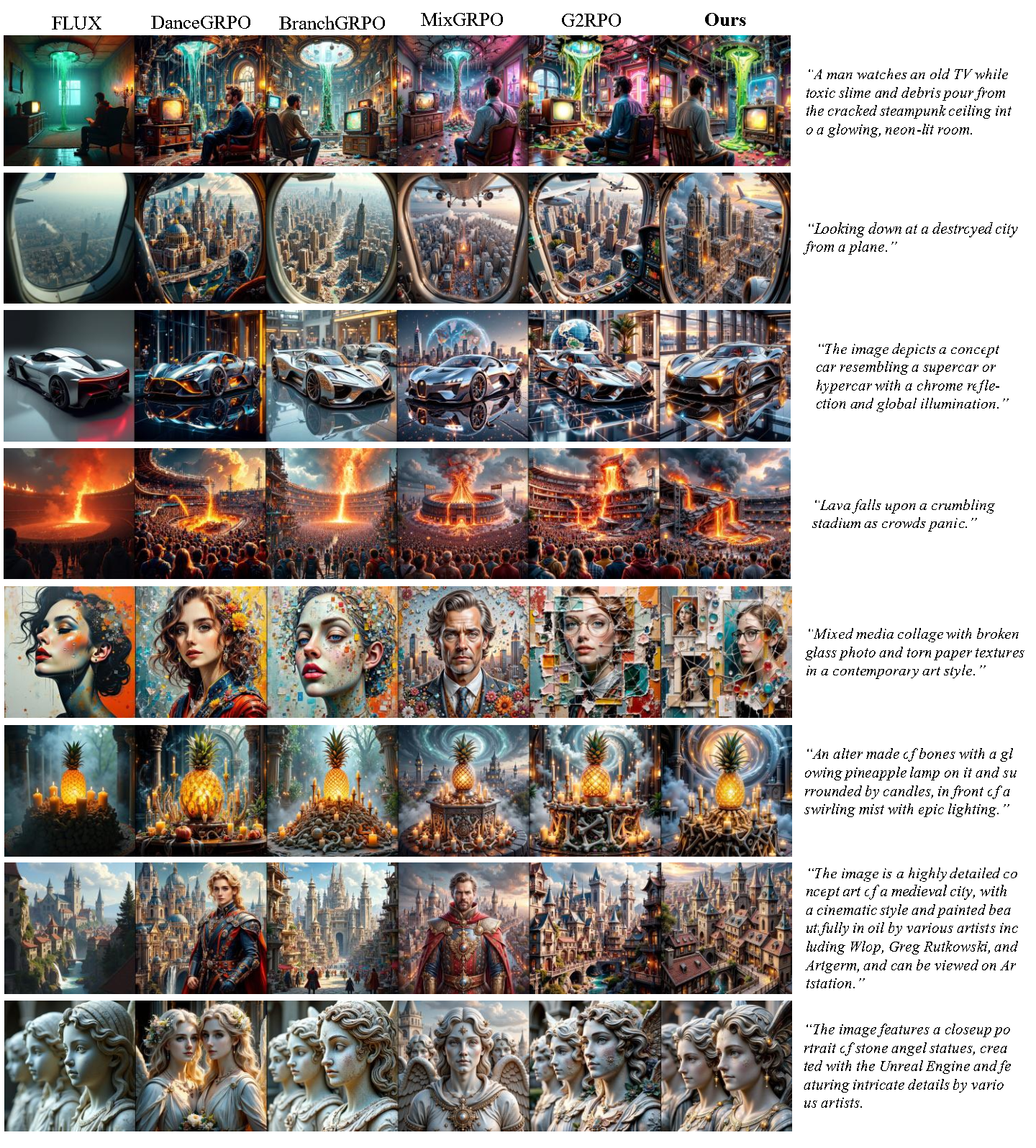} 
    \caption{\textbf{Additional visualization comparisons between E-GRPO and other baseline methods.}}
    \label{fig:vis_appendix1}
\end{figure*}

\begin{figure*}[t] 
    \centering
    \includegraphics[width=\textwidth]{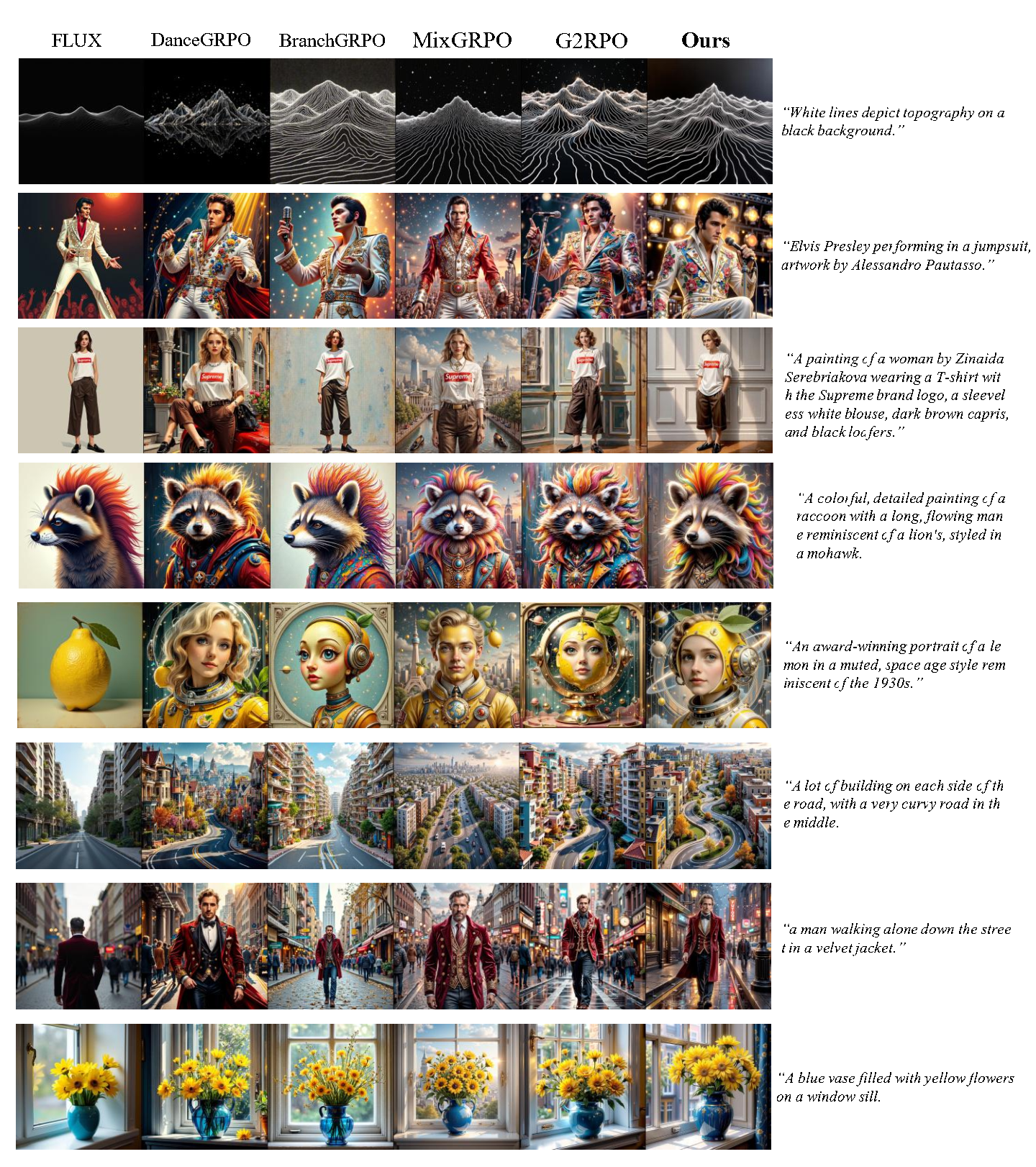} 
    \caption{\textbf{Additional visualization comparisons between E-GRPO and other baseline methods.}}
    \label{fig:vis_appendix2}
\end{figure*}

To further demonstrate the superiority of our proposed E-GRPO, we provide additional qualitative comparisons with baseline methods in Figure~\ref{fig:vis_appendix1} and Figure~\ref{fig:vis_appendix2}. As illustrated in these figures, E-GRPO consistently produces results that are more faithful to the text prompts. For example, under the prompt ``
An award-winning portrait of a lemon in a muted, space age style reminiscent of the 1930s.'' E-GRPO successfully generates a portrait that combines a space-age aesthetic with the intended compositional structure. Likewise, for the prompt ``A lot of buildings on each side of the road, with a very curvy road in the middle.'' our method captures the ``curvy'' characteristic more accurately and achieves higher aesthetic quality compared with baseline methods. These results further validate that by focusing on high-entropy steps, E-GRPO enables more effective exploration and better alignment with complex human preferences.

\subsection{Failure Cases}

Despite the robustness of E-GRPO, we observe several recurring failure patterns when handling challenging prompts.

\begin{figure*}[t] 
    \centering
    \includegraphics[width=\textwidth]{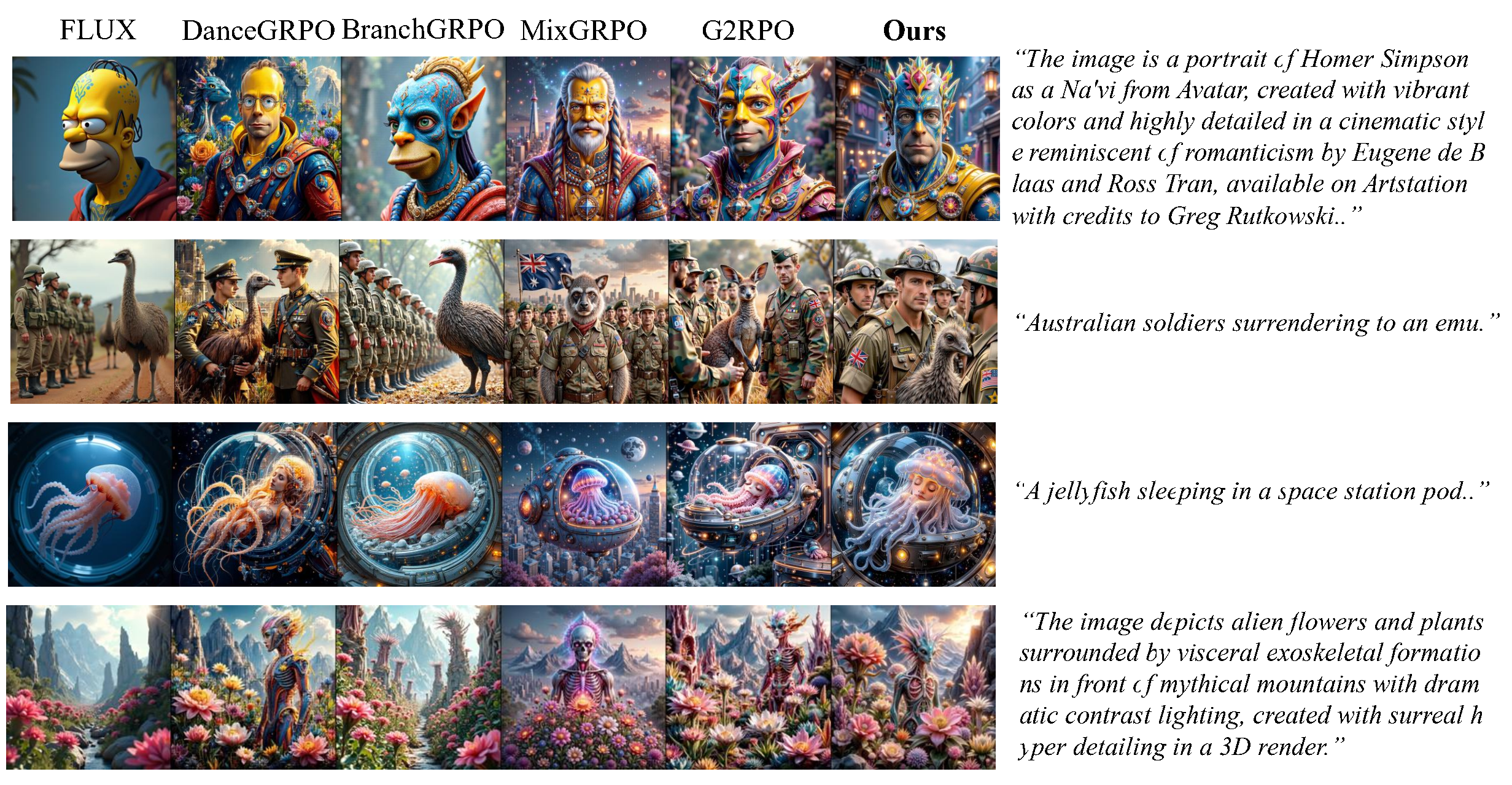} 
    \caption{\textbf{Failure cases of E-GRPO}}
    \label{fig:vis_appendix3}
\end{figure*}

\noindent \textbf{Reward Hacking.}
As discussed in the main paper, using only the HPS reward tends to produce overly saturated images, making the CLIP reward necessary as a counter-balance. Nevertheless, reward hacking still occurs in some cases. For instance, in the prompts shown in Figure~\ref{fig:vis_appendix3}, such as ``A jellyfish sleeping in a space station pod. '' and ``The image depicts alien flowers and plants surrounded by visceral exoskeletal formations in front of mythical mountains with dramatic contrast lighting, created with surreal hyper detailing in a 3D render. '', the model occasionally introduces human faces or humanoid shapes that should not be present. These artifacts reflect the model's tendency to exploit biases in the reward models, a limitation that is common across many RL-based training frameworks. Improving reward model reliability will be crucial for advancing RL in visual generation.

Overall, these observations highlight several key challenges faced by RL-based visual generation systems. Future research may explore solutions guided by these identified limitations.

\end{document}